\documentclass{article}
\usepackage[utf8]{inputenc}
\usepackage[english]{babel}
\usepackage{helvet}
\usepackage{algorithm}
\usepackage{algorithmic}
\usepackage{amsmath}
\usepackage{amssymb}

\usepackage[a4paper, total={5.9in, 9in}]{geometry}

\usepackage{graphicx}

\begin{document}

\thispagestyle{empty}
\begin{center}
    \large
%%%%%%%%%% Insert your title
    \textbf{Challenges and Responses in the Practice of Large Language Models}
    
%%%%%%%%%% Insert Author name      
    \vspace{0.4cm}
    \normalsize
    \textbf{Hongyin Zhu}
    
%%%%%%%%%% Insert Department         
    % \vspace{0.4cm}
    % \textit{Department address written in italic}
    
%%%%%%%%%% Insert your Email    
    \vspace{0.4cm}
    \textit{hongyin\_zhu@163.com}
    
    \vspace{0.4cm}
       
\end{center}
\begin{center}
\textbf{Abstract}
\end{center}
This paper carefully summarizes extensive and profound questions from all walks of life, focusing on the current high-profile AI field, covering multiple dimensions such as industry trends, academic research, technological innovation and business applications. This paper meticulously curates questions that are both thought-provoking and practically relevant, providing nuanced and insightful answers to each. To facilitate readers' understanding and reference, this paper specifically classifies and organizes these questions systematically and meticulously from the five core dimensions of computing power infrastructure, software architecture, data resources, application scenarios, and brain science. This work aims to provide readers with a comprehensive, in-depth and cutting-edge AI knowledge framework to help people from all walks of life grasp the pulse of AI development, stimulate innovative thinking, and promote industrial progress.
\begin{center}
\rule{0.9\linewidth}{0.4pt}
\end{center}
\section{Computing Power Infrastructure}
\noindent 
\textbf{Question: What is the cloud-edge-end collaborative architecture?}

The cloud-edge-end collaborative architecture is a distributed system architecture that aims to effectively integrate the computing, storage, communication, control and other resources of the cloud (the server side of the cloud service provider), the edge (the device side connected to the cloud service) and the terminal (user devices or sensors, etc.) to achieve collaborative work. This architecture integrates the resources of cloud computing, edge computing and terminal computing to achieve efficient resource scheduling and secure and reliable data transmission, thereby supporting the needs of various complex application scenarios \cite{zhu2023metaaid}, such as the Internet of Things, artificial intelligence, smart cities and industrial automation.

Specifically, the workflow of the cloud-edge-end collaborative architecture may include the following links: 1. Data collection: terminal devices and sensors are responsible for collecting various data, such as environmental parameters, user behavior, etc. 2. Edge processing: The edge device performs preliminary processing and analysis on the collected data to reduce the computing pressure on the cloud and reduce the delay of data transmission. 3. Cloud computing: The cloud server receives data from the edge, performs more in-depth analysis and calculation, and generates valuable insights and decision support. In some scenarios, the cloud is mainly used to store and manage user data. 4. Collaborative work: The cloud, edge, and terminal can achieve collaborative work and resource sharing through efficient communication protocols and data exchange mechanisms.

The advantage of the cloud-edge-end collaborative architecture is that it can make full use of various computing resources, improve the overall performance and response speed of the system, and reduce the cost and risk of data transmission. In addition, it can also support a more flexible and scalable system architecture to meet the personalized needs of different application scenarios.

\vspace{0.4cm}
\noindent 
\textbf{Question: The impact of the Information Technology Application Innovation Plan related policies on enterprises.}

The Xinchuang Plan (i.e., the Information Technology Application Innovation Plan) and related policies on domestic substitution are aimed at promoting independent innovation and development of China's information technology industry. The impact of these policies on enterprises is mainly reflected in promoting technological innovation, enhancing market competitiveness, optimizing industrial structure, and ensuring information security.

However, the implementation of the Xinchuang Plan and the domestic substitution policy also faces some challenges and difficulties. For example, domestic enterprises still have certain shortcomings and bottlenecks in key technology fields; the constraints and restrictions of foreign technical standards and market rules; and the changes in user habits and market acceptance. Therefore, when implementing these policies, it is necessary to fully consider these factors and formulate scientific and reasonable policies and measures to ensure the effectiveness and sustainability of the policies.

According to the latest data from the authoritative market research organization IDC, in 2023, China's acceleration chip market has rapidly expanded to a scale of nearly 1.4 million, among which GPU cards have dominated the market with their excellent performance, with a share of up to 85\%. The shipment volume of domestic AI chips has exceeded the 200,000 mark, accounting for about 14\% of the entire market. In 2022, the shipment volume of China's acceleration chip market was about 1.09 million, and the international giant Nvidia occupied 85\% of the market share. With the continuous advancement of technology and continuous expansion of the market, domestic AI chip brands are expected to achieve greater breakthroughs and leaps in the future.

\section{Software Architecture}

\noindent 
\textbf{Question: The necessity of having your own large language model (LLM)}

1. Improving business efficiency and accuracy: Big models have strong fitting ability and generalization performance, and can automatically complete many traditional data processing and decision-making tasks, thereby improving the efficiency and accuracy of corporate business.

2. Protecting business secrets and data privacy: With the continuous increase in data volume, the protection of data privacy and confidentiality has become increasingly important. Enterprises with private exclusive big models can better protect business secrets and data privacy, avoid the risk of data leakage and external attacks, and safeguard the core interests and competitive advantages of enterprises.

3. Customized development and use: Private exclusive big models can be customized according to the business needs and characteristics of enterprises. Development and use, so as to better meet the actual needs of enterprises. For example, in the retail field, big models can make accurate recommendations and formulate marketing strategies based on the product characteristics and consumer needs of enterprises; in the manufacturing field, big models can be intelligently planned and optimized according to the characteristics and process requirements of production lines.

4. Enhance competitiveness and innovation capabilities: Owning private exclusive big models can help enterprises enhance their competitiveness and innovation capabilities. Big models can quickly process and analyze large amounts of data, helping companies better understand market and consumer demand, and plan and seize the market in advance. At the same time, big models can also provide companies with more data insights and scientific decision-making basis, improving their strategic planning and execution capabilities.

\vspace{0.4cm}
\noindent 
\textbf{Question: When to utilize fine-tuning versus when to employ RAG}

When you need to strengthen the model's existing knowledge or adapt to complex instructions, fine-tuning \cite{zhu2021collaborative} is a good choice. Fine-tuning updates the parameters of the entire model by performing supervised learning on a labeled dataset of the new task, thereby improving the model's performance on the new task. Advantages: It can improve the model's interaction efficiency and make the model better adaptable to new tasks. Disadvantages: It consumes computing resources and training time, and is prone to overfitting problems when resources are limited or data is insufficient.

RAG is suitable for scenarios that require a lot of external knowledge, such as knowledge-intensive tasks. RAG can provide more accurate and relevant answers and enhance the interpretability of the model by combining retrievers and generators. Advantages: It can provide richer and more accurate external knowledge and enhance the model's answering ability. Disadvantages: Compared with fine-tuning, RAG has a more complex architecture, and it may be more challenging to optimize the module.

\vspace{0.4cm}
\noindent 
\textbf{Question: What were the key challenges encountered during the training of LLMs?}

1. High computing resource consumption: big model training requires a lot of computing resources, including high-performance GPUs and a large amount of storage space \cite{zhu2023heterogeneous}. This may lead to high training costs and high requirements for hardware resources.

2. Hyperparameter search: The effect of big model training is directly related to the hyperparameter configuration. It is important to search for the best hyperparameters for specific data sets and application scenarios.

3. Data management: Faced with problems such as data diversity, data coverage, data noise, and data quality, big models are prone to underfitting and overfitting problems, resulting in poor performance of the model on new data and hallucinations.

4. Interpretability: The complexity and number of parameters of big models often make their decision-making process opaque, which may lead to difficulties in attribution and tracing. At the same time, this also limits the application of models in scenarios that require high interpretability.

5. Risk control: The training and use of big models may trigger a series of AI safety issues, such as bias, violation, and unfairness. If the training data contains biased, misleading, and toxic information, the model may internalize these biases and lead to corresponding results.

6. Evaluation of big model performance: Use public benchmarks for evaluation. Automated evaluation indicators can be used for some tasks, and manual evaluation can be used for some tasks.

\section{Data Resources}
\noindent 
\textbf{Question: How to annotate a supervised fine-tuning (SFT) dataset?}

1. Clarify the task and goal:
Determine the purpose and goal of the dataset, such as whether it is used for fine-tuning a language model, classification tasks, or other NLP tasks.
Determine what types of data the dataset needs to contain, such as text, images, etc.

2. Data collection:
Collect raw data from various sources (such as the Internet, internal databases, etc.).
Ensure the diversity and representativeness of the dataset to cover a variety of possible scenarios and situations.

3. Data cleaning:
Preprocess the collected data, including removing noise, standardizing the format, etc.

4. Annotation specification formulation:
Develop detailed annotation specifications to clarify the meaning and annotation standards of each label.
Ensure the consistency and accuracy of the annotation specifications so that consistency can be maintained between different annotators.

5. Annotate data:
Annotate the data according to the annotation specifications. This can be done through crowdsourcing platforms, internal teams, or professional annotation companies.

6. Quality Control:
Implement quality control steps such as cross-checking and reviewing annotation results to ensure the accuracy and quality of annotations.
Provide training and guidance to annotators to improve the quality of annotations.

7. Dataset Division:
Divide the annotated dataset into training set, validation set, and test set for model training and evaluation.

\vspace{0.4cm}
\noindent 
\textbf{Question: Standards and regulations governing the issuance of tasks on crowdsourcing platforms}

When issuing labeling tasks on crowdsourcing platforms, you may indeed encounter the problem of poorly defined standards and specifications. This is usually due to the complexity of the task itself and the subjectivity of the labeler. To solve this problem, the following measures can be taken:

1. Develop detailed labeling guidelines:
Provide clear and specific labeling guidelines to clarify the meaning and labeling standards of each label.
Use examples and case studies to help labelers understand the labeling specifications.

2. Trial labeling and review:
Ask labelers to conduct trial labeling and review their labeling results to assess their accuracy and consistency.
Provide training and guidance to labelers who do not meet the requirements, or reassign tasks to other labelers.

3. Regular feedback and updates:
Regularly collect feedback and questions from labelers, and update the labeling guidelines and specifications based on actual conditions.
Summarize and answer common problems in the labeling process for labelers' reference.

\vspace{0.4cm}
\noindent 
\textbf{Question: When constructing a knowledge graph question-answering dataset, does it pose an issue of neglecting vital dimensions of the knowledge graph?}

When creating a knowledge graph question-answering dataset, it is a challenge to ensure that the questions are diverse and comprehensive enough to cover all important dimensions of the knowledge graph. Here are some strategies and suggestions that can help solve the problem that annotators may miss some knowledge graph dimensions when asking questions:

1. Clear knowledge graph structure:
Before starting annotation, deeply understand and analyze the structure of the knowledge graph, and clarify the key entities, attributes, relationships and their importance.
Develop a detailed annotation guide that clearly lists all dimensions that need to be covered, as well as example questions and possible answer patterns for each dimension.

2. Design diverse question templates:
According to the different dimensions of the knowledge graph, design multiple types of question templates, including asking about the basic attributes of entities, relationship queries, logical reasoning, etc.
Ensure that the question templates can cover the main aspects of the knowledge graph while avoiding duplication and redundancy.

3. Stage-by-stage annotation and review:
Carry out the annotation task in stages, with each stage focusing on different dimensions or fields of the knowledge graph.
Establish an audit mechanism for experienced annotators or experts to review the annotated data to ensure the comprehensiveness and accuracy of the questions.

4. Feedback and Iteration:
Encourage annotators to discuss and share experiences with each other, identify and improve missing dimensions.
Based on the audit results and feedback, regularly update the annotation guidelines and question templates to continuously optimize the annotation process.

5. Automated Assistance Tools:
Use natural language processing (NLP) and machine learning technologies to develop automated tools to identify dimensions that may be missed during the annotation process.
For example, a prototype of an automatic question-answering system based on knowledge graphs can be developed to assist annotators in discovering potential missing dimensions by generating questions.

6. Community Participation:
Invite experts, researchers, and community members in the field of knowledge graphs to participate in the annotation process, and use their expertise and experience to supplement and improve the dataset.
Promote cross-domain cooperation and exchanges through workshops, seminars and other activities, and jointly improve the quality of the dataset.

7. Continuous Maintenance and Update:
Recognize that knowledge graphs are dynamic and regularly update the dataset to reflect the latest changes in the knowledge graph.
Encourage user feedback and data sharing to promptly discover and correct errors and omissions in the dataset.

8. Quality Assessment and Assurance:
Implement a strict quality assessment mechanism to ensure the accuracy and reliability of the dataset.
Use multiple assessment methods, such as manual assessment, automated testing, and cross-validation, to comprehensively assess the quality of the dataset.

\vspace{0.4cm}
\noindent 
\textbf{Question: What challenges arise when utilizing LLMs for evaluating returned results?}

When using a LLM \cite{zhu2024climatechangelargelanguage} to evaluate the returned results, if the evaluation process is limited to semantic considerations, there is indeed a tendency to challenge the limitations of the model through carefully designed examples, thereby intentionally exposing the Model deficiencies in certain aspects may even amplify these specific problems. This strategy is often used for model robustness testing or performance boundary exploration, aiming to identify and optimize model weaknesses.

On the other hand, the diversity and complexity of user input may also significantly affect the performance of LLM. Different users may express similar needs in different ways, or the input may contain noise, ambiguity, or incompletely accurate information, which may lead to unsatisfactory LLM evaluation results.

In order to deal with these problems, we can adopt the following strategies to improve and optimize:

1. Build a comprehensive evaluation system:
Design assessment cases that include multiple types, styles and difficulties to comprehensively examine LLM's semantic understanding, logical reasoning, context grasp and other abilities.
Introduce a combination of manual assessment and automated assessment to ensure the objectivity and accuracy of assessment results.

2. Enhance the generalization ability of the model:
During the model training phase, focus on improving its adaptability and robustness to different types of data, by increasing the diversity and complexity of training data.

3. Optimize user input processing:
Develop an intelligent pre-processing module to perform automatic error correction, semantic analysis and intent recognition on user input to reduce model performance degradation caused by user input problems.
Provide user guidance or feedback mechanisms to help users express their needs in a more effective way, thereby improving the assessment accuracy of LLM.

4. Continuous iteration and optimization:
Based on evaluation results and user feedback, LLM is continuously iterated and optimized to improve its accuracy in processing complex inputs and evaluation results.

\section{Application Scenarios}

\noindent 
\textbf{Question: What is the mechanism behind Gemini Live, and can it be implemented through engineering practices?}

Gemini Live is a new voice chat function launched by Google, and its working principle is similar to that of GPT-4o. Users can choose multiple voices to communicate in the conversation, achieving a seamless conversation experience. Gemini Live pays special attention to the free flow of conversation, allowing users to interrupt while the other party is speaking. This design allows users to interrupt or pause at any time in the conversation, which is very suitable for scenarios that require multitasking. Even when the phone is locked, Gemini Live can work in the background to ensure that users can get information at any time.

The engineering implementation of Gemini Live involves multiple technical fields. By representing multimodal inputs as sequence tokens for processing, the input modules are different, and the unified representation module in the middle can be shared. We can draw inspiration from the architectures of llava and Qwen-audio. The input does not require OCR text recognition tools or speech recognition tools, and can achieve end-to-end understanding output. They process the input signal through ViT and audio encoding modules, and the subsequent decoders can be based on the llama3 model.

\vspace{0.4cm}
\noindent 
\textbf{Question: What challenges arise when extracting specific data tables from documents, and how can they be overcome?}

In document management, accurately locating the location of multiple tables and their pages is the first step, which is crucial for subsequent data processing and analysis. Faced with the complex and changeable table structure in the document, especially those without boxes or special layouts, it is undoubtedly a challenge to accurately parse and convert them into standard CSV format. At this time, tools such as Camelot have become the leader among many solutions with their efficient and accurate table content extraction capabilities.

However, with the advancement of technology, more and more studies have explored the use of multimodal big models to directly understand and parse tables in documents. This method has shown great potential in complex scenarios and can more intelligently capture the semantics and structural information of tables. Nevertheless, optimizing the document processing process from the source, that is, presenting and submitting the table data in a structured Json format separately during the document preparation stage, is undoubtedly the best practice to improve data processing efficiency and accuracy. This approach not only simplifies the subsequent data extraction and conversion work, but also ensures the consistency and reusability of the data, laying a solid foundation for data analysis and mining.

\vspace{0.4cm}
\noindent 
\textbf{Question: How is GraphRAG utilized and what are its key features compared to RAG }

GraphRAG is a RAG (Retrieval Augmented Generation) system that combines knowledge graphs \cite{zhu2016research} and LLMs. It significantly improves the accuracy and scalability of RAG systems by leveraging graph relationships to discover and verify information. GraphRAG has applications in many fields, such as question answering, information retrieval, etc. It generates more accurate and comprehensive answers by reasoning and validating data.

The knowledge that the RAG model relies on is often discrete and fragmented. In contrast, knowledge graphs organized through graph structures exhibit highly systematic and structured characteristics. Knowledge graph uses graph structure as the storage basis. This design not only promotes the effective integration of knowledge, but also greatly facilitates the knowledge discovery process, which is embodied in a series of advanced functions such as graph storage, graph query, graph search and graph computing. Once constructed, the knowledge graph forms a self-consistent and comprehensive knowledge system that transcends the boundaries of traditional data storage forms \cite{zhu2023pre}, whether it is structured data in databases, semi-structured information in JSON files, or widely distributed Knowledge fragments in unstructured text, as well as multi-modal data, are seamlessly integrated and transformed into interactive graph structures. The advantage of this integration is that users no longer need to care about where the knowledge is stored. They only need to query the graph through natural language or graph query language, and they can freely explore and dig into the required information, which greatly improves knowledge acquisition. efficiency and convenience. Therefore, the knowledge graph is not only an innovation in the way of knowledge representation and storage, but also an important cornerstone for promoting intelligent applications to a higher level.

\vspace{0.4cm}
\noindent 
\textbf{Question: In an enterprise environment, is there a situation where only document data needs to be processed without building a complex knowledge graph? Is knowledge graph the preferred way to organize this data only when faced with diversified, heterogeneous and multimodal data from the Internet? In addition, is it natural to recommend the use of knowledge graphs in all scenarios because of the focus on knowledge graph research, rather than considering it based on specific needs?}

First, regarding the data processing needs of users in the enterprise environment, some users do need to process documents without building complex knowledge graphs. This is because document processing usually involves basic operations such as reading, editing, storage, and retrieval, which can be well supported in the existing RAG without the need to introduce more complex knowledge graph technology.

Second, when faced with multi-source, heterogeneous, and multimodal data from the Internet, knowledge graphs have become a very effective way to organize data. Knowledge graphs can integrate these complex data sources and clearly represent the relationship between entities through graph structures, thereby helping users better understand and analyze data. This ability is particularly important when processing large-scale and complex data sets.

Although knowledge graphs have many advantages, they are not suitable for all scenarios. When choosing whether to use knowledge graphs, we need to make a comprehensive assessment based on specific needs, data characteristics, and processing complexity. Only when it is determined that knowledge graphs can bring significant benefits should we consider using them. After adopting knowledge graphs, its precision will be significantly improved and recall will be reduced. GraphRAG can effectively solve the problem of understanding knowledge graphs by combining the advantages of knowledge graphs and LLMs. It uses knowledge graphs as a structured repository of factual information, and uses LLMs for reasoning and generation, thereby achieving accurate answers to complex queries. In addition, GraphRAG also supports the combination of multimodal features and can process multiple types of data such as text and images.

Therefore, in an enterprise environment, some users may only need to process document materials without building a knowledge graph; and when faced with diversified, heterogeneous and multimodal data on the Internet, knowledge graphs may become the preferred way to organize data. However, whether to use knowledge graphs still needs to be weighed and selected according to specific needs.

\vspace{0.4cm}
\noindent 
\textbf{Question: In the news domain, how can the issue of LLMs recognizing 'USA' and 'America' as the same entity be resolved?}

After the big model identifies the entity, solving the problem of whether USA and America are the same entity mainly involves entity disambiguation and entity linking technologies. The big model only solves one link in the entire requirement. Entity disambiguation refers to solving the problem of polysemy of a word with the same name, such as Apple. Link the identified entity with the entity in the knowledge base (such as Wikipedia, DBpedia, etc.). Determine whether they represent the same entity by calculating the similarity between the entity and the entity in the knowledge base (such as vector-based similarity calculation). For example, the entity page and redirect page information in Wikipedia can be used to confirm whether USA and America are linked to the same page. After confirming that USA and America are the same entity, entity normalization is required, that is, all different names pointing to the entity are unified into a standard form (such as "United States of America"). This helps with subsequent entity association, retrieval, and data analysis.

\vspace{0.4cm}
\noindent 
\textbf{Question: In the realm of software security, how can knowledge graph technology be leveraged to achieve entity alignment across vulnerability databases? What are the advantages and disadvantages of this approach when compared to big model matching methods?}

In the field of software security, knowledge graph technology \cite{zhu2024node} builds a knowledge network of vulnerability databases with rich information and clear structure through deep structured data and precise relationship mining. It defines key entities such as vulnerabilities, software, and manufacturers, as well as complex relationships such as "impact" and "repair". After graphical organization, an intuitive and dynamic graph is formed, with nodes representing entities and edges representing relationships, forming a clear information network. The entity alignment (aka. entity matching, entity resolution) model solves the problems of data redundancy and inconsistency, ensures the uniqueness and accuracy of entities, and improves database availability. This not only accelerates security risk assessment, but also provides a solid data foundation for vulnerability repair and emergency response.

Advantages:
1. Structured representation: Knowledge graphs represent knowledge in a structured form, making the relationship between entities clearer, more intuitive, and easier to understand and query.
2. Strong interpretability: Compared with the black box characteristics of big models, the alignment process of knowledge graphs is more transparent, and the alignment results can be explained by analyzing entities and relationships.
3. Domain adaptability: In the field of software security, knowledge graphs can make full use of domain expertise and rules to improve the accuracy and pertinence of alignment.
4. Low data dependence: The alignment process of knowledge graphs mainly depends on the structure and relationship of the data itself, and has a low degree of dependence on external training data.

Disadvantages:
1. High construction cost: Building a knowledge graph requires a lot of manpower and time to define entities, relationships, and rules, and to preprocess and clean the data.
2. Poor flexibility: The structure of the knowledge graph is relatively fixed, and it is difficult to quickly adapt to changes and updates in data. In contrast, big models can adapt to new data and tasks through retraining.
3. Dependence on domain knowledge: The construction and alignment process of the knowledge graph requires the participation and guidance of domain experts to ensure accuracy and reliability. This limits its popularity and application scope to a certain extent.

\vspace{0.4cm}
\noindent 
\textbf{Question: In the field of robotics, does the integration of robots with big models possess significant practical application value?}

In the field of robots, the combination of robots and big models has shown extremely broad and practical application value. This combination not only strengthens the robot's perception and cognitive capabilities, but also greatly improves its multi-modal perception capabilities, making it capable of complex and changeable multi-task scenarios.

Take housekeeping robots as an example, such as advanced systems such as Aloha. They need to handle a series of trivial and detailed tasks from sweeping the floor and folding quilts to cooking and watering flowers. Each task requires the robot to possess different professional knowledge and skills, which places extremely high demands on the robot's intelligence level. By introducing big models into the design of domestic robots, we can achieve the following significant advantages:

1. Enhanced perception capabilities: The big model can process and analyze data from a variety of sensors, including vision, hearing, touch, etc., thereby giving the robot more comprehensive and accurate perception capabilities. This improvement in multi-modal perception enables robots to better understand and adapt to complex changes in the home environment.

2. Optimized cognitive capabilities: big models have powerful learning and reasoning \cite{tiwari2021dapath} capabilities, and can perform knowledge learning and pattern recognition based on massive data. This enables the housekeeping robot to quickly call upon relevant knowledge and formulate and execute reasonable action plans when faced with different tasks. At the same time, big models can also help robots coordinate and optimize tasks to ensure overall work efficiency and effectiveness.

3. Flexible task processing capabilities: With the support of big models, housekeeping robots can handle various tasks more flexibly. Whether it is simple sweeping the floor, folding quilts, or complex tasks such as cooking and watering flowers, the robot can make intelligent judgments and decisions based on the current environment and user needs. In addition, big models can help robots continuously learn and optimize their skills to adapt to changing household needs.

4. Improved user experience: The combination of big models and housekeeping robots not only improves the robot's work efficiency and accuracy, but also greatly improves the user experience. Users can interact with the robot through natural language, giving instructions or making demands. The robot can accurately understand the user's intention and give corresponding feedback and execution results. This intelligent interaction method makes housekeeping robots an indispensable assistant and partner in family life.

\vspace{0.4cm}
\noindent 
\textbf{Question: What scenarios are best suited for the long-context language model and RAG, and what are their respective advantages and disadvantages?}

Long-context language models are particularly suitable for scenarios that require processing large amounts of continuous text and understanding long-range dependencies. For example, fields such as legal research, medical diagnosis, and financial analysis usually require in-depth understanding and analysis of long documents.

Advantages:
1. Long-range dependency understanding: It can preserve long-range dependencies between texts, thereby more accurately understanding the information in long documents. 2. Improve information processing capabilities: As the context window grows, the model can quickly search and retrieve information from a large amount of data, improving research efficiency and data analysis. Disadvantages:
1. High computing resource consumption: Processing long texts requires more computing resources and memory, and has high hardware requirements. 2. High training difficulty: The training process of long texts is complex, requiring longer training time and larger data sets.

RAG is suitable for scenarios that require combining a large amount of external knowledge to generate answers, such as question-answering systems, content creation, etc. It can enhance the model's answering ability by retrieving external knowledge bases.

Advantages:
1. Knowledge richness: Ability to retrieve external knowledge bases, provide more comprehensive and in-depth information, and enhance the quality of the model's answers.
2. High accuracy: Combining the retrieved information, it can generate more accurate answers and reduce the "hallucination" problem that may occur in the generative model.
3. Flexibility: The RAG framework allows the model to adapt to a variety of tasks and has high flexibility.
Disadvantages:
1. Dependence on external data: Need to rely on external knowledge bases. If the knowledge base is not comprehensive or updated in time, it may affect the quality of the answer.
2. Retrieval efficiency: The retrieval process may increase the response time of the system and affect the user experience.
3. Integration complexity: The retrieval system and the generative model need to be effectively integrated, which may be complex to implement, increasing the complexity and maintenance cost of the system.

\vspace{0.4cm}
\noindent 
\textbf{Question: What are the key technological differences in the stacks employed by various types of AI search, including Perplexity AI, Big model-powered search, AI-powered search solutions from traditional search companies, and AI search startups?}

Currently, there are significant differences in technology stacks for different types of AI searches. These differences are mainly reflected in basic models, technology integration, application scenarios and optimization strategies.

1. Perplexity AI is built upon cutting-edge LLMs, including GPT-3.5 and GPT-4, as well as the Bing search engine API, which have powerful language understanding and generation capabilities.
Technology integration: Perplexity combines generative AI with search technology, and combines it with Microsoft's Bing search engine to create a new AI-empowered conversational search engine.
Application scenario: Provide direct answers and summaries, and cite relevant sources to support multi-turn conversations and contextual memory capabilities.
Optimization strategy: Optimize product performance through continuous iteration, improve user experience, and launch a variety of tool products such as Copilot, Bird SQL, etc.

2. Big model-powered search's primary focus is on serving as the backbone of a multi-turn dialogue robot. Additionally, it supports the uploading of documents for RAG.
Technology integration: Deeply integrate big models with vector databases to alleviate the hallucination problem.
Application scenario: Chatbot and Knowledge base Q\&A.
Optimization strategy: They emphasize the importance of continuously refining and optimizing their foundational models. This includes enhancing the model's ability to process long contexts, accelerating inference speeds, and implementing other improvements that directly impact model performance and accuracy.

3. For AI-powered search solutions from traditional search companies, traditional search companies usually optimize based on their own accumulated massive data and search technology, combined with generative AI technology.
Technology integration: Integrate AI technology on the basis of the original search engine to improve the intelligence and personalization level of search and improve the effect of result display.
Application scenarios: Continue to consolidate and expand the search engine market and provide richer and more personalized search results.
Optimization strategy: Combining user feedback and data analysis to continuously optimize search algorithms and user experience, while exploring new business models and growth points.

4. AI search startups frequently opt for either open-source or commercialized big models as their foundational platform. They then integrate with search engine APIs, meticulously fine-tuning and optimizing these models to cater to the unique requirements of local markets and individual user preferences. 
Technology integration: AI search startups integrate advanced technologies such as semantic analysis and knowledge graphs into their foundational models. These integrations enable a deeper understanding of user queries, allowing for more relevant and accurate search results. 
Application scenarios: You can focus on specific domain or scenarios, such as e-commerce search, academic search, legal search, etc., or you can search in general domain.
Optimization strategy: Continuously improve product performance and user experience through technological innovation and iterative optimization, while exploring cooperation opportunities with upstream and downstream enterprises.

\vspace{0.4cm}
\noindent 
\textbf{Question: What is the relationship between graph computing, graph neural networks, and knowledge graphs? In the field of graph computing, do you think that more emphasis should be placed on research on graph database storage rather than graph neural networks? In addition, I would like to know what is the purpose of representing nodes as vectors in graph neural networks?}

Graph computing, graph neural networks (GNNs), and knowledge graphs are closely related in AI and big data, but have distinct research focuses and applications.

Graph computing focuses on processing and analyzing graph-structured data, using algorithms like path search, centrality measures, and community detection. It encompasses not just graph database storage but also efficient data handling, feature extraction, and various computational tasks.

GNNs are deep learning models tailored for graph data. They capture dependencies through message passing and excel in tasks like classification, clustering, and link prediction. Their ability to handle non-Euclidean data makes them useful in social networks, recommendation systems, and bioinformatics. GNN research focuses on designing effective neural architectures and leveraging complex patterns for prediction and reasoning.

Knowledge graphs represent knowledge as graphs of entities and relationships, supporting applications like QA, recommendations, and search. They provide rich graph data for GNNs, which enhance knowledge graph representation and applications through tasks like node classification, link prediction, and relation reasoning.

In summary, knowledge graphs create graph data, graph computing explores it, and GNNs leverage it to enhance representation and application capabilities.

Graph database storage is a crucial aspect of graph computing, focusing on efficient and reliable storage of graph-structured data for subsequent processing and analysis. However, graph computing research extends beyond storage to encompass data handling, feature extraction, and various computational tasks on graphs.

Node vector representations form the basis for GNNs across tasks, capturing graph data's intrinsic structure and patterns to aid prediction and reasoning. Without them, machine learning model training is unfeasible. Dense representations power deep learning, and enable efficient parallel computation, speeding up inference.

\vspace{0.4cm}
\noindent 
\textbf{Question: With the emergence of big model technologies, is the relevance of knowledge graphs diminishing or being abandoned?}

The rise of big model technology has profoundly impacted the construction and application of knowledge graphs. Despite their distinct architectures and application scopes, they complement each other, forming a symbiotic relationship. This synergy creates a new mode of dual knowledge engines (big models and knowledge graphs) working together to push the boundaries of AI development.

Big models, with their powerful representation learning, cross-domain generalization, and complex task-handling capabilities, breathe new life into AI. Trained on vast data, they capture richer, nuanced knowledge representations, addressing knowledge graphs' limitations in coverage, update speed, and reasoning flexibility.

Meanwhile, knowledge graphs, as the cornerstone of structured knowledge storage and reasoning, offer precise, explainable, and easily queryable knowledge representations crucial for AI applications requiring high accuracy and interpretability. Their entities, relationships, and attributes provide a solid semantic foundation for intelligent services, enabling complex queries, reasoning, and decision support, mitigating big models' hallucinations and logical errors.

Nowadays, knowledge graph projects are no longer isolated but deeply integrated with big model technology, forming a new paradigm of dual knowledge engines working in harmony. This collaboration enhances knowledge integration, utilization, and overall intelligence capabilities, such as understanding, reasoning, and creativity. big models can learn from knowledge graphs to improve domain-specific performance, while knowledge graphs can update and expand with new insights from big models.

Once a research hotspot, knowledge graphs have shown immense potential in information retrieval, question answering, and recommendation systems. With big model technology's growth, knowledge graphs' application scenarios have expanded further, jointly driving AI innovation.

In summary, the fusion of big model technology and knowledge graphs is an inevitable trend and a key to AI's leapfrog development. As technologies mature and applications deepen, the dual knowledge engine model will play a pivotal role in various fields, fostering AI's comprehensive development and widespread adoption.

\section{Brain Science}

\noindent 
\textbf{Question: What is the current progress and trajectory of the industrial transformation within the field of brain science?}

The industrial transformation of brain science \cite{zhu2020species} is on an accelerated track and has made remarkable milestones. On the one hand, the commercialization process of brain-computer interface technology is quietly emerging. It has revolutionized the seamless connection between the human brain and advanced external devices, opening up an unprecedented path for the instant transmission and fine control of information. This technology not only heralds great potential in improving the quality of life of patients, but also heralds that the medical field is about to usher in a new era of personalized and precise treatment, bringing hope to countless patients.

On the other hand, the fruitful results of brain science \cite{poo2016china} research are profoundly affecting the development trajectory of the field of artificial intelligence. By integrating the profound insights of brain science into the research and development of AI technology, it not only gives artificial intelligence systems capabilities that are closer to human thinking, but also greatly promotes the expansion of the boundaries and performance leaps of AI technology. This interdisciplinary integration not only provides a solid theoretical foundation and source of inspiration for the technological innovation of the AI industry, but also paves the way for the infinite possibilities of future intelligent technology.

More importantly, brain science plays an irreplaceable role in protecting human brain health and conquering brain diseases \cite{zhu2016brain}. It not only provides scientific basis and technical support for the early diagnosis and precise treatment of brain diseases, but also helps to build a more comprehensive and systematic brain health management system, building a solid line of defense for human health and well-being.

In summary, the industrial transformation of brain science is not only a scientific and technological revolution, but also an important contribution to the quality of human life and future development. With its unique charm and unlimited potential, it leads us towards a new era of greater intelligence and health.

\vspace{0.4cm}
\noindent
\textbf{Question: What valuable insights can the field of brain science offer to inform the future development and advancement of Transformer models?}

The profound inspiration of brain science to the Transformer model is specifically reflected in the following dimensions, showing the wonderful resonance between the two in information processing and cognitive functions:

1. Attention mechanism: The self-attention mechanism in the Transformer model is a simplified simulation of the brain's efficient information processing strategy. When faced with complex information, the brain can quickly lock on to key information and ignore redundant details, a highly selective attention allocation mechanism.

2. Memory mechanism: The human brain has a complex and sophisticated memory system, including short-term memory and long-term memory, as well as an efficient memory storage and retrieval mechanism. This biological characteristic provides valuable inspiration for the model architecture in memory processing, by drawing on the brain's memory mechanism.

3. Multi-brain region collaborative information processing paradigm: Human cognitive functions do not exist in isolation, but rely on close collaboration and information exchange between multiple brain regions to form brain circuits (brainconnectome, brain connectivity) for cognitive functions. This multi-brain region collaborative mechanism helps the model's design ideas when building complex information processing systems. By simulating the functional division of labor and synergy of different brain regions, more complex cognitive tasks can be achieved.

4. Dynamic system perspective of brain-inspired mechanism: As a highly dynamic system, the internal mechanism of the brain is far beyond the scope of simple electrical signal transmission. Complex phenomena such as the formation and forgetting of memories, the fluctuation and regulation of emotions, etc. often involve complex reactions and regulation of chemical substances. This perspective prompts us to not only focus on the optimization of the computational level when designing the Transformer model, but also explore how to introduce more diversified mechanisms (such as dynamic weight adjustment, emotional computing, etc.) to build a brain-inspired model that is closer to human intelligence.

5. Energy consumption:
The energy consumption of the brain is much lower than that of the Transformer big model, mainly because:
Efficient biological components: The energy efficiency of neurons and synapses far exceeds that of electronic components. 
Parallel and distributed processing: Information processing is highly parallel and distributed, which improves efficiency and reduces energy consumption. 
Sparse connections: The connections between neurons are sparse, which reduces unnecessary information processing.
Adaptive plasticity: It can optimize neural networks based on learning and experience to reduce energy consumption. 
Evolutionary optimization: Long-term evolution has enabled the brain to develop efficient energy consumption mechanisms. 
Efficient energy utilization: Relying on glucose for energy, without energy reserves, emphasizing the importance of efficient energy consumption.

\vspace{0.4cm}
\noindent
\textbf{Question: Can the design and functionality of agents' memory systems be inspired and informed by advancements in brain science?}

The memory of agents can indeed be inspired by brain science. The brain's memory mechanisms include short-term memory, long-term memory, working memory, etc. These mechanisms provide important references for the memory design of agents. For example, inspired by the brain's working memory mechanism, the DNC (Differential Neural Computer) in artificial neural networks divides sequence control and memory storage into two modules, improving the ability to handle complex tasks. In addition, the continual learning mechanism in the brain also provides inspiration for agents to learn another new skill based on mastering one skill.

% \printbibliography
\bibliographystyle{unsrt} % 指定参考文献的样式  
\bibliography{Literature} % 指定.bib文件名，不需要.bib扩展名 

\begin{thebibliography}{10}

\bibitem{zhu2023metaaid}
Hongyin Zhu.
\newblock Metaaid 2.5: A secure framework for developing metaverse applications
  via large language models.
\newblock {\em arXiv preprint arXiv:2312.14480}, 2023.

\bibitem{zhu2021collaborative}
Hongyin Zhu, Prayag Tiwari, Ahmed Ghoneim, and M~Shamim Hossain.
\newblock A collaborative ai-enabled pretrained language model for aiot domain
  question answering.
\newblock {\em IEEE Transactions on Industrial Informatics}, 18(5):3387--3396,
  2021.

\bibitem{zhu2023heterogeneous}
Hongyin Zhu, Ke~Liu, and Zhongxiang Sun.
\newblock Learning commonsense knowledge graph embedding via entity semantic
  aggregation.
\newblock In {\em Proceedings of CCKS}, 2023.

\bibitem{zhu2024climatechangelargelanguage}
Hongyin Zhu and Prayag Tiwari.
\newblock Climate change from large language models, 2024.

\bibitem{zhu2016research}
Hongyin Zhu, Yi~Zeng, and Yiping Yang.
\newblock Research topics variation analysis and prediction based on faro and
  neural networks.
\newblock In {\em 2016 IEEE International Conference on Systems, Man, and
  Cybernetics (SMC)}, pages 000910--000915. IEEE, 2016.

\bibitem{zhu2023pre}
Hongyin Zhu, Hao Peng, Zhiheng Lyu, Lei Hou, Juanzi Li, and Jinghui Xiao.
\newblock Pre-training language model incorporating domain-specific
  heterogeneous knowledge into a unified representation.
\newblock {\em Expert Systems with Applications}, 215:119369, 2023.

\bibitem{zhu2024node}
Hongyin Zhu.
\newblock Node classification via semantic-structural attention-enhanced graph
  convolutional networks.
\newblock {\em arXiv preprint arXiv:2403.16033}, 2024.

\bibitem{tiwari2021dapath}
Prayag Tiwari, Hongyin Zhu, and Hari~Mohan Pandey.
\newblock Dapath: Distance-aware knowledge graph reasoning based on deep
  reinforcement learning.
\newblock {\em Neural Networks}, 135:1--12, 2021.

\bibitem{zhu2020species}
Hongyin Zhu, Yi~Zeng, Dongsheng Wang, and Cunqing Huangfu.
\newblock Species classification for neuroscience literature based on span of
  interest using sequence-to-sequence learning model.
\newblock {\em Frontiers in Human Neuroscience}, 14:128, 2020.

\bibitem{poo2016china}
Mu-ming Poo, Jiu-lin Du, Nancy~Y Ip, Zhi-Qi Xiong, Bo~Xu, and Tieniu Tan.
\newblock China brain project: basic neuroscience, brain diseases, and
  brain-inspired computing.
\newblock {\em Neuron}, 92(3):591--596, 2016.

\bibitem{zhu2016brain}
Hongyin Zhu, Yi~Zeng, Dongsheng Wang, and Bo~Xu.
\newblock Brain knowledge graph analysis based on complex network theory.
\newblock In {\em Brain Informatics and Health: International Conference, BIH
  2016, Omaha, NE, USA, October 13-16, 2016 Proceedings}, pages 211--220.
  Springer, 2016.

\end{thebibliography}
\end{document}